%% file: main.tex
\documentclass[10pt,twocolumn,letterpaper]{article}

\usepackage[pagenumbers]{cvpr} 

\input{preamble}

\definecolor{cvprblue}{rgb}{0.21,0.49,0.74}
\usepackage[pagebackref,breaklinks,colorlinks,urlcolor=cvprblue,linkcolor=cvprblue,citecolor=cvprblue]{hyperref}
\usepackage{inconsolata}
\usepackage{multirow}
\usepackage{tabularx}
\usepackage{colortbl}
\usepackage{pifont}
\usepackage{multicol}
\usepackage{xcolor}

\usepackage{colortbl}
\usepackage{xcolor}
\definecolor{tabhighlight}{HTML}{e5e5e5}
\renewcommand{\ttdefault}{zi4}
\definecolor{mydarkgreen}{rgb}{0.0, 0.5, 0.0}  

\usepackage{color}
\usepackage{marvosym}
\usepackage{enumitem}
\usepackage{utfsym}
\usepackage{xspace}
\usepackage{pifont}
\usepackage{arydshln}
\usepackage[T1]{fontenc}
\definecolor{citecolor}{HTML}{0071BC}
\definecolor{linkcolor}{HTML}{ED1C24}
\definecolor{LGray}{gray}{0.97}

\usepackage{wrapfig}

\usepackage{sidecap}
\usepackage{booktabs}
\renewcommand{\arraystretch}{1}
\usepackage{afterpage}

\title{PG-Video-LLaVA: Pixel Grounding Large Video-Language Models}
\author{Shehan Munasinghe$^{1*}$, Rusiru Thushara$^{1*}$, Muhammad Maaz$^{1}$, Hanoona Abdul Rasheed$^{1}$, \\ Salman Khan$^{1,2}$, Mubarak Shah$^{4}$, Fahad Khan$^{1,3}$\\
$^{1}$Mohamed bin Zayed University of AI, $^{2}$Australian National University 
\\ $^{3}$Linköping University,
$^{4}$University of Central Florida \\[0.5em]
Project: {\tt  \href{https://github.com/mbzuai-oryx/Video-LLaVA}{https://github.com/mbzuai-oryx/Video-LLaVA} }
}

\begin{document}
\maketitle
\def \thefootnote{*}\footnotetext{Equal Contribution}
\input{sec/0_abstract}    
\input{sec/1_intro}
\input{sec/2_related_works}
\input{sec/3_method}

\input{sec/4_experiments}
\input{sec/5_conclusion}

{
    \small
    \bibliographystyle{ieeenat_fullname}
    \bibliography{main}
}

\input{sec/X_suppl}

\end{document}

%% file: preamble.tex
\usepackage[dvipsnames]{xcolor}


%% file: sec/0_abstract.tex
\begin{abstract}
Extending image-based Large Multimodal Models (LMMs) to videos is challenging due to the inherent complexity of video data. The recent approaches extending image-based LMMs to videos either lack the grounding capabilities (e.g., VideoChat, Video-ChatGPT, Video-LLaMA) or do not utilize the audio-signals for better video understanding (e.g., Video-ChatGPT). Addressing these gaps, we propose PG-Video-LLaVA, the first LMM with pixel-level grounding capability, integrating audio cues by transcribing them into text to enrich video-context understanding. Our framework uses an off-the-shelf tracker and a novel grounding module, enabling it to spatially localize objects in videos following user instructions. We evaluate PG-Video-LLaVA using video-based generative and question-answering benchmarks and introduce new benchmarks specifically designed to measure prompt-based object grounding performance in videos. Further, we propose the use of Vicuna over GPT-3.5, as utilized in Video-ChatGPT, for video-based conversation benchmarking, ensuring reproducibility of results which is a concern with the proprietary nature of GPT-3.5. Our framework builds on SoTA image-based LLaVA model and extends its advantages to the video domain, delivering promising gains on video-based conversation and grounding tasks. 

\end{abstract}

%% file: sec/1_intro.tex
\section{Introduction}
\label{sec:intro}

Recent efforts on Large Multimodal Models (LMMs), spearheaded by GPT-4V \cite{GPT4V}, allow detailed conversations about images but generally do not scale well to videos. The magnitude of video data scales far beyond other modalities due to its massive volume on social and internet media. Furthermore, extending LMMs to videos is challenging due to their complex dynamics with long temporal context that needs to be understood accurately. 
Although recent approaches towards video-LMMs such as VideoChat~\cite{2023videochat}, Video-LLaMA~\cite{damonlpsg2023videollama}, and Video-ChatGPT~\cite{Maaz2023VideoChatGPT} have demonstrated capabilities 
in video comprehension and dialogue, they lack the crucial feature of visual grounding. Visual grounding in videos aims to associate the LMM responses to specific objects within the video input. Addressing this gap, we introduce PG-Video-LLaVA, the first video-LMM capable of localizing objects appearing in LMM responses. This task leads to enhanced intractability and demonstrates deep understanding of video content.


\begin{figure}
  \centering
    \includegraphics[width=0.5\textwidth]{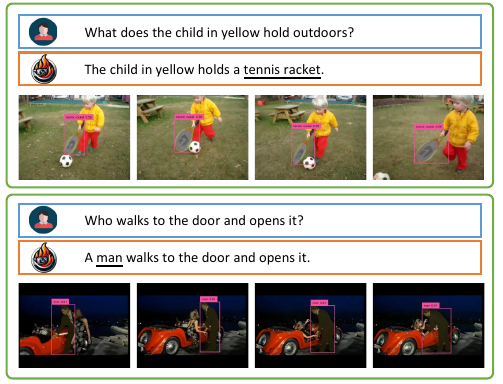}
  \caption{\textbf{Video spatial grounding} on example videos from Vid-STG~\cite{zhang2020does} (above) and HC-STVG~\cite{hcstvg} (below) datasets. PG-Video-LLaVA can generate textual responses with referred objects grounded in the video content (\emph{tennis racket} and \emph{man} are localized in the top and bottom examples, respectively). }
  \label{fig:teaset}
\vspace{-1em}
\end{figure}

In PG-Video-LLaVA, we address the unique challenges posed by video data. The model is designed to track objects within shorter video clips that maintain consistent camera views, enabling accurate visual grounding across scenes and motions. This tracking links spatio-temporal segments directly to conversational elements, enhancing the model's contextual understanding. A salient feature of PG-Video-LLaVA is its modular design, allowing for easy integration with existing grounding modules and the flexibility to adapt to future enhancements in visual grounding technology.
Moreover, PG-Video-LLaVA enriches its capabilities by incorporating audio context. It achieves this by leveraging video audio in a form understandable to LLM, which is particularly useful in situations where the auditory information is essential to the conversation. This inclusion broadens the model's understanding, making it more versatile in interpreting video content. 

Furthermore, this work introduces an improved framework for benchmarking video-based conversational models, pivoting from previous approaches~\cite{Maaz2023VideoChatGPT} that predominantly used the proprietary GPT-3.5-Turbo model for evaluation. Given that GPT-3.5-Turbo is subject to changes at any time and lacks transparency due to its closed-source nature, it presents challenges in terms of reliability and reproducibility. To address this, we propose the use of Vicuna, an open-source LLM for benchmarking. This shift not only enhances reproducibility but also improves transparency in the evaluation process. We evaluate PG-Video-LLaVA using our improved benchmarks and show notable improvements over existing video conversational models like Video-ChatGPT~\cite{Maaz2023VideoChatGPT} and Video-LLaMA~\cite{damonlpsg2023videollama} in ungrounded dialogues, achieving state-of-the-art (SoTA) performance.

The key contributions of this work are:
\begin{itemize}
    \item We propose PG-Video-LLaVA, the first video-based LMM with pixel-level grounding capabilities, featuring a modular design for enhanced flexibility.
    \item By incorporating audio context, PG-Video-LLaVA significantly enhances its understanding of video content, making it more comprehensive and aptly suited for scenarios where the audio signal is crucial for video understanding (e.g., dialogues and conversations, news videos, etc.).
    \item We introduce improved quantitative benchmarks for video-based conversational models. Our benchmarks utilize open-source Vicuna LLM to ensure better reproducibility and transparency. We also propose benchmarks to evaluate the grounding capabilities of video-based conversational models.
\end{itemize}


%% file: sec/2_related_works.tex
\section{Related Works}
\label{sec:related_works}


Recent advancements in Large Multimodal Models (LMMs) \cite{Liu2023VisualIT,zhu2023minigpt,Dai2023InstructionBLIP} and Large Language Models (LLMs) \cite{chiang2023vicuna, OpenAI2023ChatGPT, touvron2023llama} have significantly transformed the artificial intelligence landscape, particularly in natural language processing and multimodal tasks. These breakthroughs have enhanced machine learning models' ability to understand and generate human-like text, while also enabling more effective integration of various data types like images, sounds and videos with textual information. This progress represents a major leap in creating AI systems that can accurately interpret and interact with a diverse range of content.

\noindent
\textbf{Large Language Models (LLMs):}
The natural language processing (NLP) field has undergone a revolution with the advent of LLMs such as GPT~\cite{brown2020language}, LLaMA~\cite{touvron2023llama}, OPT~\cite{zhang2022opt}, and MOSS~\cite{OpenLMLab2023MOSS}, particularly noted for their zero-shot learning abilities and adaptability. The development of models like InstructGPT~\cite{ouyang2022training} and ChatGPT~\cite{OpenAI2023ChatGPT} has further propelled advancements in conversational AI and complex query handling, chiefly through instruction tuning. Within the LLaMA framework, the emergence of open-source models such as Alpaca~\cite{taori2023stanford} and Vicuna~\cite{chiang2023vicuna} exemplifies how instruction tuning can significantly boost model performance. This shift towards open-source initiatives in language modeling, highlighted by models like Alpaca and Vicuna, indicates a growing trend towards more accessible and collaborative approaches in the field. In this work, we build on the open-source Vicuna LLM and extend it with multimodal capabilities. We also propose an open-source benchmark for video conversation and reasoning tasks using Vicuna LLM that is reproducible for fair evaluations.

\noindent
\textbf{Large Multimodal Models (LMMs):}
The field of AI has witnessed significant advancements with the development of vision-language models like CLIP~\cite{radford2021learning}, renowned for their impressive zero-shot capabilities using extensive image-text pairs during training. These models have proven effective in a variety of applications, from image detection and segmentation~\cite{liang2023open, bangalath2022bridging} to more complex tasks such as 3D modeling and video analysis~\cite{rozenberszki2022language, ni2022expanding, wang2021actionclip, rasheed2023fine}. The introduction of BLIP-2 marked a pivotal transition, pioneering the integration of image features encoded by a visual encoder with text embeddings, setting the stage for the evolution into Large Multimodal Models (LMMs). This advancement influenced subsequent models like LLaVA~\cite{liu2023visual}, InstructBLIP~\cite{Dai2023InstructionBLIP}, and MiniGPT-4~\cite{zhu2023minigpt}, which further refined image-text feature alignment and instruction tuning. VideoChat~\cite{2023videochat}, Video-ChatGPT~\cite{Maaz2023VideoChatGPT} and Video-LLaMA~\cite{damonlpsg2023videollama} represents an extension of these LMMs, moving from image-based to video-based applications, while models such as Otter~\cite{li2023otter}, mPLUG-Owl~\cite{ye2023mplug}, LLaMa-Adapter~\cite{gao2023llamaadapterv2}, and InternGPT~\cite{liu2023internchat} continue to push the boundaries of multimodal interaction. Despite these significant strides, challenges in achieving robust visual grounding in LMMs highlight key areas for ongoing research and development in this dynamic field. Further,  effective integration of audio signals within LMMs for comprehensive video understanding is an open research question that this work aims to address.

\noindent
\textbf{Visual-Language Grounding:}
Grounded Large Language Models (LLMs) have made notable progress in enhancing visual and language comprehension. A diverse array of models including Kosmos-2~\cite{peng2023kosmos}, Ferret~\cite{you2023ferret}, All-Seeing Model~\cite{wang2023all}, LISA~\cite{lai2023lisa}, BuboGPT~\cite{zhao2023bubogpt}, Shikra~\cite{chen2023shikra}, and GLaMM~\cite{rasheed2023glamm} have employed various methodologies to master complex grounding tasks. These models demonstrate proficiency in tasks like referring expression comprehension and image segmentation, showcasing the advanced image understanding capabilities of LLMs. Methodologically, Kosmos-2, Shikra, and All-Seeing focus predominantly on creating language-based context for visual grounding. In contrast, BuboGPT merges visual elements with language, and LISA leverages vision-language embeddings for producing segmentation masks. Furthermore, GLaMM is adept at generating natural language responses linked with object segmentation masks, facilitating detailed visual-textual interactions. However, challenges remain, such as LISA's constrained performance in multi-object scenarios and the limitations of BuboGPT and GLaMM to image-based applications, not extending to video processing. To this end, we introduce PG-Video-LLaVA, a video conversational model with pixel-level grounding capability. Further, PG-Video-LLaVA incorporates audio transcripts alongside visual and textual data, aiming to provide a more detailed understanding of video content.

%% file: sec/3_method.tex
\section{PG-Video-LLaVA}
\label{sec:method}

\begin{figure*}[htp]
   \centering
   \includegraphics[width=0.98\textwidth]{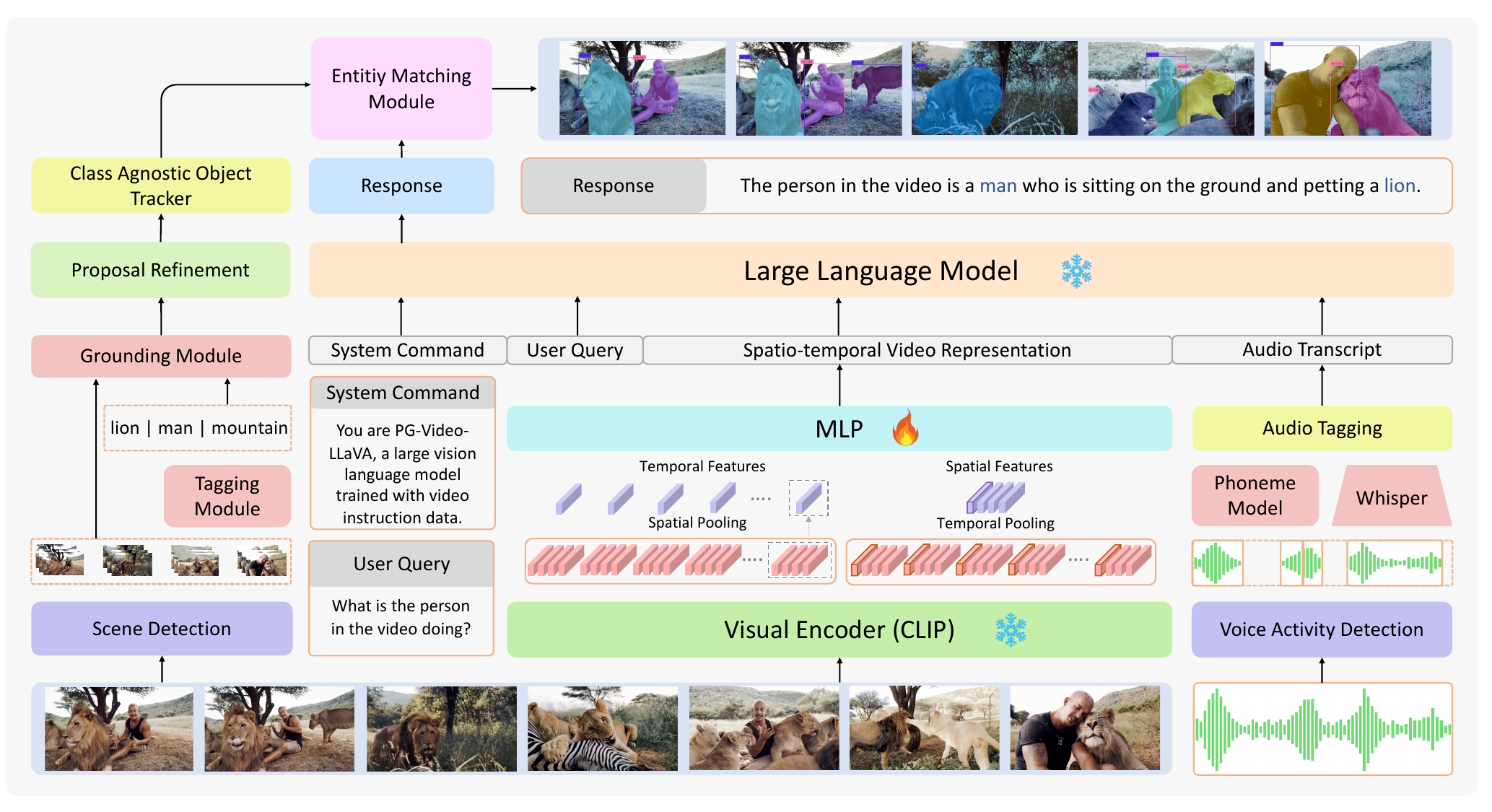}
   \vspace{-0.5em}
   \caption{\textbf{Architecture of PG-Video-LLaVA: } PG-Video-LLaVA integrates a CLIP-based visual encoder with a multimodal language model for video understanding. The CLIP visual encoder extracts spatio-temporal features from videos by averaging frame-level features across temporal and spatial dimensions. These features are then projected into the LLM's input space using a learnable Multi-Layer Perceptron (MLP). The system features a grounding module for spatially locating textual descriptions within video frames, a class-agnostic object tracker, and an entity-matching module. Audio processing incorporates voice activity detection, phoneme modeling, and Whisper-based audio transcription, resulting in a multimodal pipeline that facilitates robust video-question answering. The architecture is trained on a hybrid dataset of video instructions, enabling the handling of diverse conversational contexts with high accuracy.}
   \label{fig:overall-architecture}
\end{figure*}



\subsection{Overview}

In this paper, we introduce PG-Video-LLaVA, a novel Large Multimodal Model (LMM) designed to align video and audio representations with a Large Language Model (LLM). This integration equips PG-Video-LLaVA with the capability to proficiently manage both video and audio data in conversational contexts. Additionally, our method integrates a specialized plug-and-play module for effective video grounding (see Figure~\ref{fig:overall-architecture}).

In constructing PG-Video-LLaVA, our approach integrates sophisticated mechanisms for aligning video and audio signals with language processing capabilities, thereby facilitating a comprehensive multimodal analysis. Central to our model is an advanced CLIP-based video encoder, which has been specifically adapted to process both spatial and temporal dimensions of video data. This adaptation enables a deeper understanding of video content, setting PG-Video-LLaVA apart from conventional image-centric models.


For training, PG-Video-LLaVA utilizes the VideoInstruct100K \cite{Maaz2023VideoChatGPT} dataset comprising 100K video instructions derived from ActivityNet-200 \cite{Heilbron2015ActivityNetAL}. This diverse dataset ensures that the model is well-equipped to handle a broad spectrum of video contexts with high accuracy. In addition to visual processing, PG-Video-LLaVA incorporates state-of-the-art audio analysis by leveraging advanced audio transcription techniques, similar to those employed in WhisperX \cite{bain2023whisperx} and Whisper-AT\cite{gong_whisperat}. This integration allows the model to process and understand audio inputs effectively, enhancing its overall multimodal interpretation capabilities.

While PG-Video-LLaVA's foundation is based on the LLaVA-1.5~\cite{Liu2023VisualIT} framework, it is extended for videos to incorporate spatio-temporal representations, audio understanding and visual grounding capabilities. Its unique combination of enhanced video encoding, extensive training dataset, integrated audio processing and grounding capability marks it as a forward step in the field of LMMs.

\subsection{Architecture}
Our architecture utilizes the CLIP ViT-L/14@336 as the visual encoder, which, unlike its original image-focused design, has been adapted for video processing in PG-Video-LLaVA. This adaptation is crucial for the model to capture spatio-temporal representations in videos effectively. In our model, video samples are represented as \( V_i \in \mathbb{R}^{T \times H \times W \times C} \), where \( T \) denotes the frame count. The encoder processes each of the \( T \) frames independently, treating them as a series of images. This leads to the generation of frame-level embeddings \( x_i \in \mathbb{R}^{T \times h \times w \times D} \), where \( h = H/p \) and \( w = W/p \), with \( p \) being the patch size (14 for ViT-L/14) and \( N = h \times w \) indicating the total token count.

To construct a comprehensive video-level representation, we apply average pooling across the temporal dimension of these frame-level embeddings, resulting in a temporal representation \( t_i \in \mathbb{R}^{N \times D} \). This temporal pooling technique effectively amalgamates information across multiple frames. Similarly, for spatial information, we achieve spatial representation \( z_i \in \mathbb{R}^{T \times D} \) through average pooling along the spatial dimension. The final video-level features \( v_i \) are a combination of these temporal and spatial features, as shown in the equation:

\begin{equation}
v_i = [t_i \quad z_i] \in \mathbb{R}^{(T+N)\times D}
\end{equation}

In our architectural design, the spatio-temporal feature extraction is inspired by Video-ChatGPT~\cite{Maaz2023VideoChatGPT}, with an additional enhancement of employing a higher resolution of 336×336 pixels to encode frame-level features.

Within the architecture of PG-Video-LLaVA, we have implemented a learnable Multi-Layer Perceptron (MLP), designated as \( g \), to serve as our cross-modal connector. This MLP is intricately designed to project video-level features into the embedding space of the language decoder. This is inspired from LLaVA-1.5 \cite{Liu2023VisualIT}, aiming to optimize the model's multi-modal capabilities beyond what could be achieved with a simple linear projection. The process yields language embedding tokens \( Q_v \), calculated as follows:

\begin{equation}
Q_v = g(v_i) \in \mathbb{R}^{(T+N)\times K}
\end{equation}

Text queries, denoted as \( Q_t \in \mathbb{R}^{L \times K} \) where \( L \) is the length of the query, are tokenized to be dimensionally compatible with these video embeddings. The combination of \( Q_v \) and \( Q_t \) is then fed into the language decoder, facilitating the seamless integration of video and textual data within the model (see Figure~\ref{fig:overall-architecture}).




\subsubsection{Audio Modality Integration}
In PG-Video-LLaVA, we have integrated an audio processing pipeline that significantly enhances the video-question answering capabilities by incorporating audio cues from the input, drawing inspiration from the architecture of WhisperX\cite{bain2023whisperx}. The process begins with the deployment of a Voice Activity Detection (VAD) model. This model is crucial for pinpointing speech-containing temporal segments within the audio track. Following the VAD's identification of speech segments, these segments undergo processing—cutting, merging, and padding—to align with the input specifications of the Whisper model \cite{WhisperOpenAI}. Simultaneously, a phoneme segmentation model operates in parallel, producing phone-level segmentations essential for the subsequent alignment of raw transcriptions with the audio.

The VAD model serves a dual purpose: it not only identifies speech segments but also aids in filtering out non-speech audio components. To enhance the compatibility of transcriptions generated by Whisper with our model, we integrate Whisper-AT\cite{gong_whisperat}. This advanced version of the Whisper model specializes in audio tagging. It annotates the audio stream with labels from an extensive set of 527 audio event classes, allowing for precise temporal resolution.

The audio transcripts are then subjected to a multi-stage filtering process. Initially, a VAD-based filter is applied, followed by a phoneme-based forced alignment using the Whisper model, ensuring temporally accurate text transcriptions. Utilizing Whisper's language identification feature, we eliminate non-English speech segments at this stage. For each identified sentence segment, we apply Whisper-AT \cite{gong_whisperat} for audio tagging, focusing on the top three predicted audio classes. Segments that do not predominantly feature `speech', or where `music' probabilities significantly exceed `speech', are excluded from further processing.

Finally, the integration of the audio transcript with the video component is executed through a carefully designed prompt template. This template is pivotal in guiding the system to understand user instructions, assimilate the video frames, and incorporate the transcriptions generated by the automatic speech recognition model. This structured approach ensures that PG-Video-LLaVA efficiently leverages all available modalities—visual and auditory—thereby enabling users to achieve task completion and query resolution based on a comprehensive analysis of both visual and auditory content (refer to Figure~\ref{fig:overall-architecture} for details).

\label{sec:experiments}
\begin{table*}[t]
\centering
\resizebox{2\columnwidth}{!}{
\begin{tabular}{lccccc}
\toprule
\multirow{2}{*}{\textbf{Model}} & \multicolumn{5}{c}{\textbf{Evaluation Metrics}} \\
\cmidrule{2-6}
 & \textbf{Correctness} & \textbf{Detail Orientation} & \textbf{Contextual Understanding} & \textbf{Temporal Understanding} & \textbf{Consistency} \\
\midrule
LLaMA Adapter~\cite{gao2023llamaadapterv2} & 2.34  & 2.43  & 2.65 & 2.20  & 3.02 \\
Video Chat~\cite{2023videochat} & 2.49  & 2.82 & 2.92 & 2.27 & 3.11  \\
Video-LLaMA~\cite{damonlpsg2023videollama} & 2.29 & 2.61 &	2.68 &	2.17 &	2.87  \\
Video-ChatGPT~\cite{Maaz2023VideoChatGPT} & 2.51 &	2.53 &	2.85 &	2.32 & 3.10 \\
\rowcolor{violet!10} PG-Video-LLaVA (7B) &  2.73 & 2.89 & 3.13 & 2.44 & 3.40  \\
\rowcolor{violet!10} PG-Video-LLaVA (13B) &  \textbf{2.86} & \textbf{2.95} & \textbf{3.23} & \textbf{2.53} & \textbf{3.49}  \\
\bottomrule
\end{tabular}
}
\caption{\textbf{Performance benchmarking of video-based conversational models.} Comparative performance evaluation of PG-Video-LLaVA against various models using the benchmarking framework from Video-ChatGPT~\cite{Maaz2023VideoChatGPT}. The metrics include correctness, detail orientation, contextual understanding, temporal understanding, and consistency. The updated assessment pipeline incorporates Vicuna-13b-v1.5~\cite{chiang2023vicuna} for enhanced reproducibility, replacing GPT-3.5-Turbo. Results indicate that PG-Video-LLaVA achieves favourable performance across all metrics, particularly in contextual and temporal understanding, as compared to foundational models and recent advancements in the field.}
\label{tab:table1}
\end{table*}

\subsubsection{Grounding Module}

In PG-Video-LLaVA, our spatial grounding approach starts with processing video-question pairs to generate textual descriptions. These descriptions are then used for grounding within the video frames. Key noun phrases are extracted from the generated text using Vicuna, targeting the most critical content aspects. Simultaneously, an image tagging model, RAM \cite{zhang2023recognize}, tags visual elements in each frame, creating a detailed map of the video content.


The video is segmented into smaller parts using PySceneDetect \cite{PySceneDetect}, based on changes in scene composition. This segmentation facilitates a more focused grounding process. In each segment, our grounding ensemble, composed of GroundingDINO \cite{liu2023grounding}, DEVA \cite{cheng2023tracking}, and SAM \cite{kirillov2023segany}, utilizes the image tags to create segmentation masks and tracking IDs for the identified visual elements.

The visual cues from these segmentation masks are then matched with the textual noun phrases using CLIP \cite{radford2021learning}. This matching process links text to the corresponding visual elements in the video, enhancing our understanding of the content.


In quantitative analysis, from the descriptive textual response to an interrogative text, a referring expression or a phrase is extracted using Vicuna. This phrase is input into our grounding module, which then generates segmentation masks and tracking IDs. We measure the spatial grounding accuracy of our model by calculating the Intersection over Union (IoU) between these segmentation masks and ground truth bounding boxes.

This systematic approach enables PG-Video-LLaVA to effectively ground textual descriptions within video content, thereby improving the performance and interpretability of video-question answering systems.

%% file: sec/4_experiments.tex
\begin{figure*}[t]
   \centering
   \includegraphics[width=\textwidth]{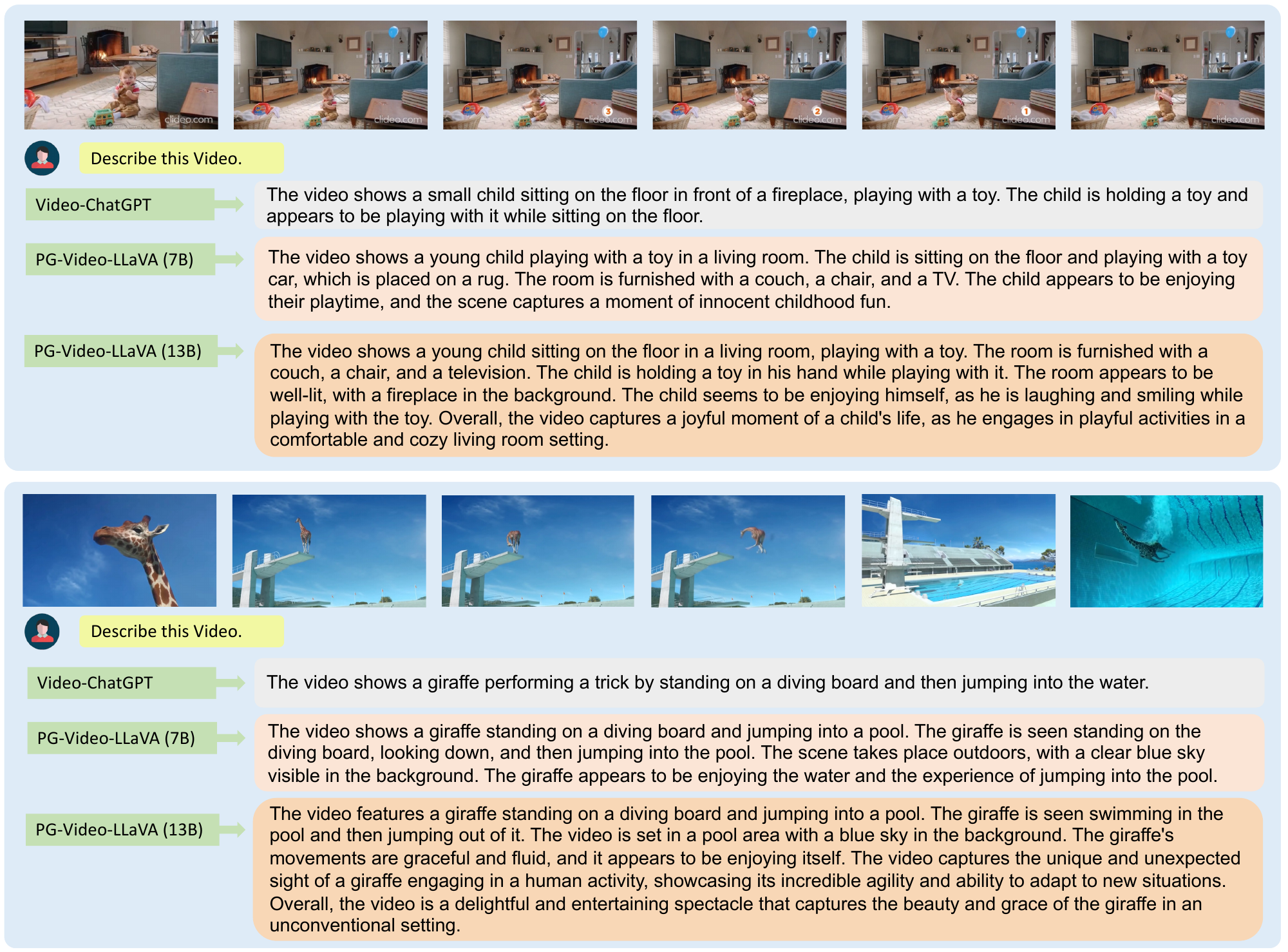}
   \caption{\textbf{Qualitative results comparison of Video-ChatGPT vs PG-Video-LLaVA (Ours)} Qualitative analysis of video descriptions generated by Video-ChatGPT, PG-Video-LLaVA (7B), and PG-Video-LLaVA (13B) models. The evolution in model performance is evident, with enhancements in the accuracy of information, richness of descriptive detail, and alignment with the video’s context and sequence of events as we move from the baseline Video-ChatGPT to the more advanced PG-Video-LLaVA (13B) model.}
   \label{fig:qualitative-comparative}
\end{figure*}

\section{Experiments}
\subsection{Implementation Details}
We build our stronger baseline on top of LLaVA-1.5 which utilizes CLIP ViT-L/14@336 as the image encoder and Vicuna 1.5 as the LLM. We only
tune the MLP projection layers during training, while keeping the rest of the architecture frozen. We finetune the model for 3 epochs using a learning rate of 2$e^{-5}$ and an overall batch size of 32. The training of our 7B and 13B models took around 6 hours and 15 hours respectively on 4 A100 80GB GPUs.

For audio transcript extraction, base Whisper model is used. Our grounding module is based on GroundingDINO-T variant and CLIP ViT-B/32. For the image-tagging model we use RAM Swin-Large variant (with input size 384). DEVA Tracker is applied under online-setting in our experiments. 

Vicuna-13b-v1.5 model is used in performing video-based conversational benchmarking, zero-shot question answering evaluation, and extracting the key noun or referring expression from the model output in the quantitative evaluation of the spatial grounding task. Further, Vicuna-13b-v1.5 was used to implement the entity matching as in~\cite{zhao2023bubogpt}.


\subsection{Stronger Baseline}
This section provides an overview of the quantitative evaluations conducted to determine the effects of the strengthened baseline on PG-Video-LLaVA. To evaluate the impact of the enhanced baseline on PG-Video-LLaVA, we apply the benchmarking framework from Video-ChatGPT\cite{Maaz2023VideoChatGPT}. This framework measures performance on several axes critical for video-based conversational agents, including correctness of information, detail orientation, contextual understanding, temporal understanding, and consistency.

In order to facilitate a reliable and reproducible evaluation, we have updated our assessment pipeline by replacing GPT-3.5-Turbo with Vicuna-13b-v1.5. This adjustment addresses the limitations in reproducibility inherent to the closed-source nature of GPT-3.5-Turbo. Subsequently, we have re-assessed both PG-Video-LLaVA and other recent models to ensure a fair and consistent comparison. The results shown in Table~\ref{tab:table1} demonstrate that PG-Video-LLaVA outperforms the foundational Video-ChatGPT model and exhibits superior performance when compared to other recent contributions in the domain.

Following the quantitative assessment, the qualitative results in Figure~\ref{fig:qualitative-comparative} indicate the enhanced baseline's impact on PG-Video-LLaVA's performance. The PG-Video-LLaVA (13B) model exhibits improved accuracy in the information presented, a deeper level of descriptive detail, and a stronger alignment with the context and temporal progression of the videos. This advancement is particularly noticeable in the precise depiction of the child's engagement with their surroundings and the giraffe's behaviour, indicating a refined interpretation of both the activities and their settings. These qualitative insights are consistent with the quantitative results, highlighting the augmented baseline's role in advancing PG-Video-LLaVA's capacity in video understanding.

\begin{figure*}[h]
   \centering
   \includegraphics[width=0.98\textwidth]{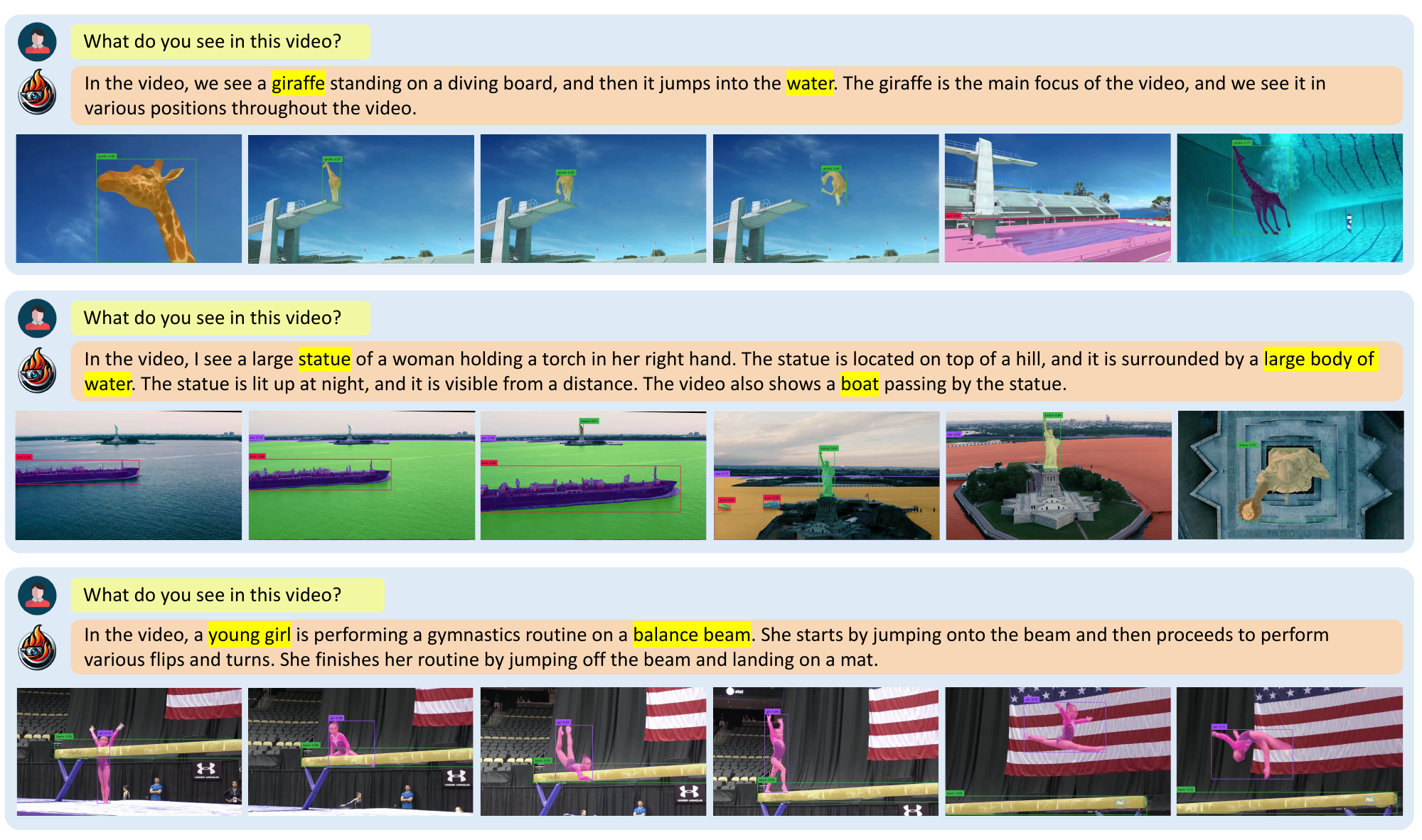}
   \caption{\textbf{Qualitative Results for Video Grounding}: Visual representation of the grounding capability of advanced video-conversational capabilities of PG-Video-LLaVA. The highlighted regions in each video frame indicate the model's ability to identify and spatially locate key subjects mentioned in the textual description, such as the giraffe, the statue, and the gymnast on a balance beam.}
   \label{fig:qualitative-grounding}
\end{figure*}

\begin{table}[!th]
\centering
\resizebox{0.95\columnwidth}{!}{
\begin{tabular}{lcc}
\toprule
\textbf{Model} & \textbf{VidSTG}~\cite{zhang2020does} & \textbf{HC-STVG}~\cite{hcstvg} \\ 
\midrule
Grounding DINO~\cite{liu2023grounding} & 25.3 & 19.5 \\
Video-LLaMA~\cite{damonlpsg2023videollama} & 28.6 & 26.1 \\
Video-ChatGPT~\cite{Maaz2023VideoChatGPT} & 32.8 & 20.8 \\
\rowcolor{violet!10} PG-Video-LLaVA (7B) & 34.2 & \textbf{28.3} \\
\rowcolor{violet!10} PG-Video-LLaVA (13B) & \textbf{35.1}  & 27.3 \\
\bottomrule
\end{tabular}
}
\caption{\textbf{Performance of PG-Video-LLaVA and other models on spatial grounding task}: Evaluated using the VidSTG and HC-STVG benchmarks, the results demonstrate PG-Video-LLaVA's favorable spatial grounding capabilities, as evidenced by its ability to generate accurate descriptive responses and effectively locate referring expressions within video frames. The table shows the model's progress, particularly in the 13B version, showcasing its performance among other SoTA video-conversational models.}
\label{results_table2}
\vspace{-1.5em}
\end{table}
\begin{table*}[!th]
\centering
\setlength{\tabcolsep}{8pt}
\renewcommand{\arraystretch}{1}
\resizebox{1.9\columnwidth}{!}{
\begin{tabular}{l c c c c c c c c}
\toprule
\textbf{Model} & \multicolumn{2}{c}{\textbf{MSVD-QA}~\cite{xu2017video_msvdqa}} & \multicolumn{2}{c}{\textbf{MSRVTT-QA}~\cite{xu2016msr-vtt}} & \multicolumn{2}{c}{\textbf{TGIF-QA}~\cite{tgif-cvpr2016}} & \multicolumn{2}{c}{\textbf{Activity Net-QA}~\cite{yu2019activityqa}} \\
\cmidrule{2-9}
 & \textbf{Accuracy} & \textbf{Score} & \textbf{Accuracy} & \textbf{Score} & \textbf{Accuracy} & \textbf{Score} & \textbf{Accuracy} & \textbf{Score} \\
\midrule
FrozenBiLM~\cite{yang2022frozenbilm} & 32.2 & -- & 16.8 & -- & 41.0 & -- & 24.7 & -- \\
LLaMA Adapter~\cite{gao2023llamaadapterv2} & 53.7 & 3.3 & 45.6 & 3.2 & 54.3 & 3.3 & 37.3 & 3.2 \\
Video LLaMA~\cite{damonlpsg2023videollama} & 48.6 & 3.2 & 32.8 & 2.8 & 51.4 & 3.4 & 27.1 & 2.9 \\
Video-ChatGPT~\cite{Maaz2023VideoChatGPT} & 62.6 & 3.6 & 50.0 & 3.3 & 66.5 & 3.7 & \textbf{40.8} & \textbf{3.3} \\
\rowcolor{violet!10} PG-Video-LLaVA  & \textbf{64.1} & \textbf{3.7} & \textbf{51.6} & \textbf{3.3} & \textbf{66.8} & \textbf{3.8} & 39.9 & \textbf{3.3} \\
\bottomrule
\end{tabular}}
\caption{\textbf{Zeroshot video-based question-answering:} Comparison of PG-Video-LLaVA with other video generative models. The latest available models are used for all the approaches and the benchmarks are calculated using open-source Vicuna LLM. PG-Video-LLaVA performs better than the previously proposed video-based conversational methods.}
\label{results_table3}
\end{table*}

\begin{figure*}[!th]
   \centering
   \includegraphics[width=0.99\textwidth]{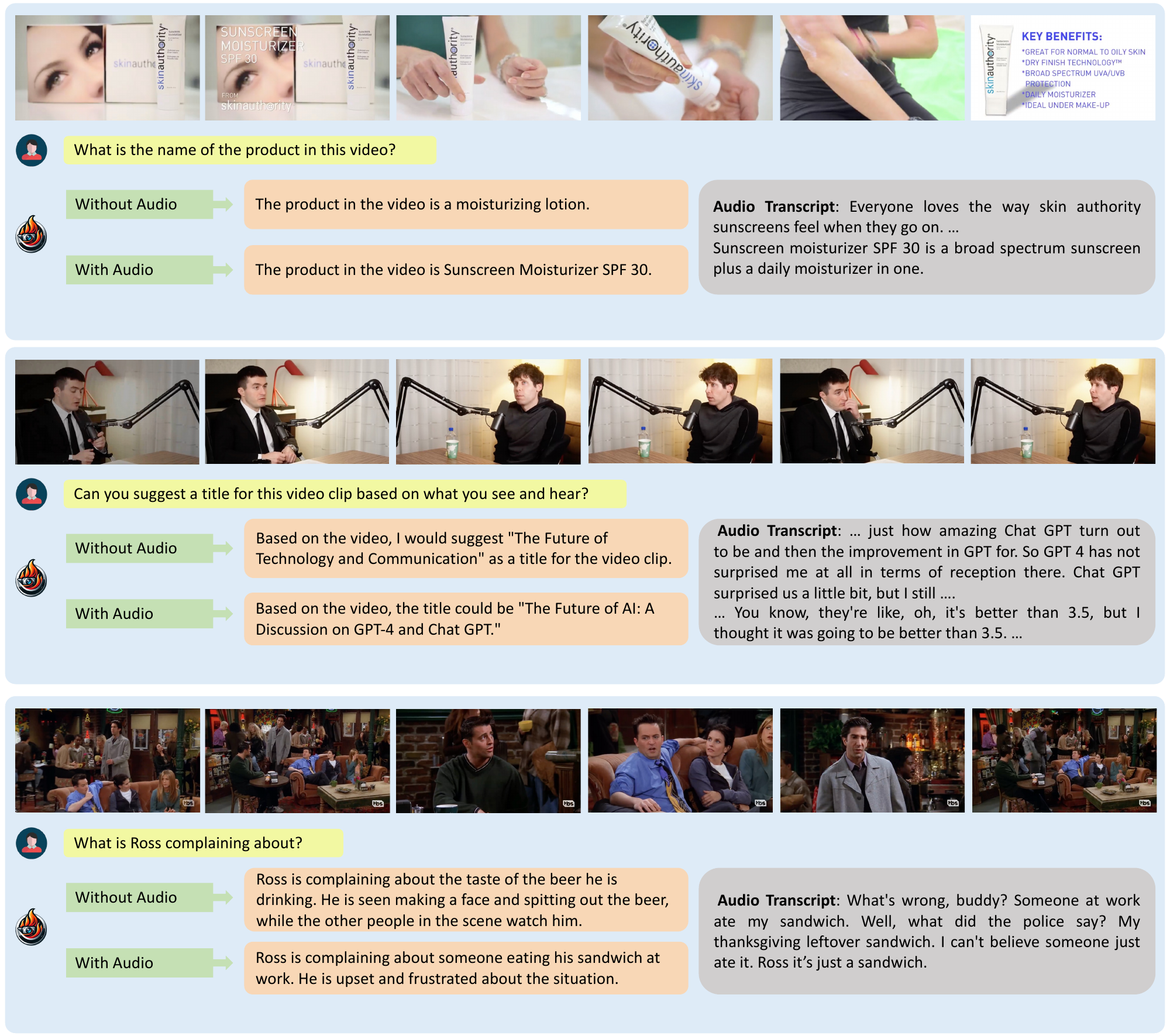}
   \caption{\textbf{Qualitative Results for Including Audio Modality}: The figure illustrates the integrated audio processing pipeline that augments video-question answering with audio cues. It provides side-by-side comparisons showing how audio cues offer additional context, leading to a more accurate interpretation of the video content, as seen in the examples above.}
   \label{fig:qualitative-audio}
\end{figure*}

\subsection{Spatial Grounding in Videos}
To quantitatively assess PG-Video-LLaVA's spatial grounding capability, we conducted quantitative evaluations of PG-Video-LLaVA's spatial grounding capabilities using two benchmarks that are derived from the test set of the VidSTG \cite{zhang2020does} and HC-STVG \cite{hcstvg} datasets. Due to the novelty of integrating spatial grounding within video-conversational models, we highlight the modular nature of our grounding pipeline, which can be incorporated with other state-of-the-art video conversation models. For the VidSTG dataset, we selectively processed interrogative prompts to assess the grounding accuracy. The model generates descriptive textual responses to these prompts, from which Vicuna-13b-v1.5 extracts relevant referring expressions. These expressions are then spatially grounded in the video frames using our grounding pipeline. For the HC-STVG dataset, interrogative prompts are first mined from the text captions using Vicuna and then used similarly to VidSTG prompts.

The results shown in Table~\ref{results_table2} position PG-Video-LLaVA alongside alternative methods using the same benchmarks, demonstrating our model's enhanced ability to accurately answer questions, thereby leading to improved spatial grounding performance.

The qualitative results shown in Figure~\ref{fig:qualitative-grounding} emphasize the model's refined spatial grounding precision. The accurate overlay of masks on the subjects within the videos confirms the model's adeptness at correlating textual descriptors with visual elements, a critical aspect of contextual comprehension. This refined ability is crucial for applications that integrate visual data with language, improving the model's utility in environments that demand rich, interactive visual and linguistic processing.

\subsection{Zero-Shot Visual Question Answering}
For PG-Video-LLaVA, zero-shot question-answering (QA) capabilities were evaluated quantitatively using several established open-ended QA datasets: MSRVTT-QA\cite{xu2016msr-vtt}, MSVD-QA \cite{xu2017video_msvdqa}, TGIF-QA \cite{tgif-cvpr2016}, and ActivityNet-QA \cite{yu2019activityqa}. These datasets are benchmarks for assessing a model's ability to generate accurate answers without any dataset-specific fine-tuning. We adopted a zero-shot evaluation methodology, utilizing Vicuna-13b-v1.5 to evaluate the model's understanding and predictive accuracy, with scores assigned on a scale from 1 to 5. Results are presented in Table~\ref{results_table3}.

In comparison to Video-ChatGPT, PG-Video-LLaVA demonstrates superior performance, surpassing not only the predecessor but also other notable models in the field, such as FrozenBiLM\cite{yang2022frozenbilm} and Video Chat\cite{2023videochat}. The results from our evaluations indicate that PG-Video-LLaVA has significantly enhanced its ability to comprehend video content and generate contextually relevant answers, thus establishing a new state-of-the-art in zero-shot VideoQA.

As shown in Figure \ref{fig:qualitative-grounding}, our method is able to visually ground the key objects in the given video. Improvement in the model's capability to describe the content in the video is demonstrated in Figure \ref{fig:qualitative-comparative}. Further, it can be observed that adding the audio modality helps make correct outputs, whereas the model without audio modality fails to capture those details from visual content alone (Figure \ref{fig:qualitative-audio}).

%% file: sec/5_conclusion.tex
\section{Conclusion}
\label{sec:conclusion}

In this work, we introduced PG-Video-LLaVA, a novel video-based conversational model equipped with pixel-level grounding capabilities. PG-Video-LLaVA enhances image-based conversational models by extracting spatio-temporal features essential for comprehensive video understanding. It incorporates filtered audio transcripts to enrich the interpretation of visual scenes where audio cues are pivotal. Additionally, we developed a novel grounding module capable of tracking and generating pixel-level grounding of objects within videos. To promote reproducibility, we propose quantitative benchmarks for video-based conversational models, utilizing the open-sourced Vicuna LLM instead of GPT-3.5, as employed by previous approaches. These benchmarks are specifically designed to evaluate grounding capabilities. In summary, this work represents the first effort to integrate grounding capabilities into video-based LMMs. 

%% file: sec/X_suppl.tex
\clearpage
\setcounter{page}{1}
\maketitlesupplementary

\appendix


\renewcommand{\ttdefault}{pcr} 

\section{Audio Modality Integration}
Here, we outline the implementation details of audio modality integration in PG-Video-LLaVA.

\subsection{Audio Transcript Filtering}

To generate audio transcripts, we first experimented with using the state-of-the-art Whisper \cite{WhisperOpenAI} directly. However, the obtained transcripts were too noisy, contained hallucinations, and unwanted text such as lyrics from songs. Passing these raw audio transcripts directly to the LLM without any filtering can negatively affect the overall model's performance. Therefore, a preprocessing method is required to filter out noisy text and keep only the parts of the audio that carry meaningful information. 

The following steps combining WhisperX\cite{bain2023whisperx} and Whisper-AT\cite{gong_whisperat} are used to refine the original Whisper transcripts to be usable as inputs to the video LMM.
\begin{enumerate}
    \item We first apply VAD-based preliminary filtering to the audio, and then use the Whisper model with Phoneme-based forced alignment to get temporally aligned text transcriptions. 
    \item As Whisper is able to identify the language spoken, all non-English speech can be ignored at this point since PG-Video-LLaVA generates responses in English.
    \item For each sentence segment obtained, slice the original audio at the corresponding timestamps and pass to Whisper-AT to produce audio-tagging output.
    \item For each sentence segment, consider the top 3 audio classes predicted.
    \begin{enumerate}
        \item If “\text{speech}” is not among the top 3 predictions, the segment is ignored.
        \item If  $P[\text{music}]>P[\text{speech}]$ and $P[\text{music}]-P[\text{speech}]>threshold$, the segment is ignored (the $threshold$ is set empirically to $1.1$).
    \end{enumerate}
\end{enumerate}

\begin{figure*}[t]
   \centering 
   \includegraphics[width=0.98\textwidth]{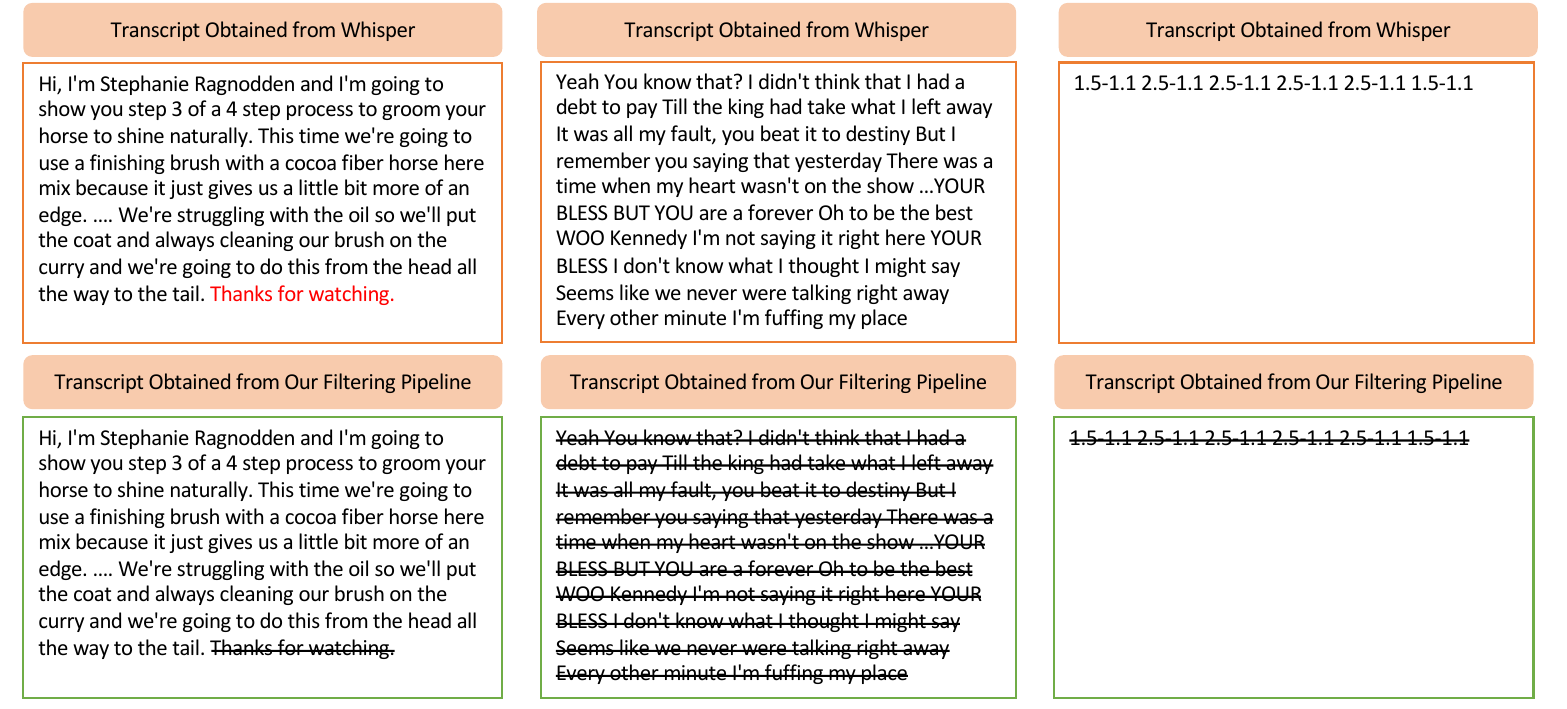} 
   \caption{\textbf{Filtering the audio transcript}: to remove hallucinations (left), music (center), and garbage (right) characters from the raw audio transcript.} 
   \label{fig:Transcript-filtering} 
\end{figure*}

Figure \ref{fig:Transcript-filtering} shows the effectiveness of our audio transcript preprocessing method in filtering out hallucinations, music, and garbage characters from the raw audio transcript.

\subsection{Integrating Audio Transcript into the LLM}

The following prompt template is used when combining the spatiotemporal video features and audio transcript with the user instruction text.

\vspace{0.5em}
\noindent \texttt{SYSTEM}: \vspace{-0.7em}
\begin{quote}
\texttt{You are PG-Video-LLaVA, a large vision-language assistant. \\
You are able to understand the video content that the user provides, and assist the user with a variety of tasks using natural language.}
\end{quote}
\noindent \texttt{USER}: \vspace{-0.7em}
\begin{quote}
\texttt{<Instruction>\\
<Video-Tokens> \\
The noisy audio transcript of this video is: <Audio-Transcript> 
}
\end{quote}
\noindent \texttt{ASSISTANT}: \vspace{-0.7em}

\vspace{0.8em}


\section{Visual Grounding: Quantitative Evaluation}

\subsection{Overview}

We introduce novel benchmarks for quantitatively evaluating conversation-based video spatial grounding, based on two existing spatio-temporal video grounding datasets, VidSTG\cite{zhang2020does} and HC-STVG\cite{hcstvg}. 

\begin{figure}[ht]
  \centering
    \includegraphics[width=0.5\textwidth]{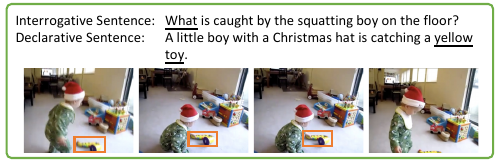}
    \vspace{-2em}
  \caption{\textbf{Interrogative vs declarative sentences}}
  \label{fig:declarative_interrogative}
\vspace{-1em}
\end{figure}

In conversation-based spatial grounding, the objective is to localize interrogative sentences with unknown objects in the given video (e.g. ``What is caught by the squatting boy on the floor?” ). Unlike grounding for declarative sentences where the explicit characteristics of objects (e.g. the class ``toy” and visual appearance ``yellow”) are present within the sentence itself, grounding for interrogative sentences is challenging due to the fact that it can only depend on relationships between the unknown object and other objects (e.g. the action relation ``caught by the squatting boy” and spatial relation ``on the floor”) (Figure \ref{fig:declarative_interrogative}). A benchmark based on this task can be regarded as a measure of the sufficient relationship construction and cross-modal relation reasoning ability of the video-language model.

To evaluate our model for conversation-based video spatial grounding, we pass interrogative prompts to the model. It then generates descriptive textual responses to these prompts, from which Vicuna-13b-v1.5 extracts relevant referring expressions. These expressions are then passed into the GroundingDINO-based spatial grounding and tracking module. For the obtained object tracks, bounding box IoU is calculated by comparing them with the ground truth annotations. 

From the two spatiotemporal grounding datasets, to form a spatial-only grounding benchmark, we crop the video in the temporal axis to contain only the segment where the target object is present, and the mean spatial IoU is reported as the metric for comparison.

It should be noted that we evaluate our model in these benchmarks only in the zero-shot setting, without any training on these datasets.

\textbf{1. Benchmark based on the VidSTG Dataset: } VidSTG dataset consists of videos paired with multiform sentences (both interrogative and declarative). To form a benchmark to quantitatively evaluate the performance of conversation-based video spatial grounding, we leverage the 5693 video and interrogative sentence pairs in its test set.

\textbf{2. Benchmark based on HC-STVG Dataset: }
Unlike in VidSTG dataset, in HC-STVG dataset contains only declarative form sentences for all of its videos. Therefore interrogative sentences are first generated from the declarative text captions in 3025 samples of the test set using Vicuna-13b-v1.5 model. Then the evaluation is performed in a similar manner to VidSTG.

\subsection{Generating Interrogative Statements}

The original text annotations in the HC-STVG dataset are in the declarative statement format. In order to make our video prompt-based grounding evaluation pipeline, we extract interrogative statements (questions) from these text annotations using Vicuna-13b-v1.5 using the following prompt template. 

\vspace{0.5em}
\noindent \texttt{SYSTEM}: \vspace{-0.7em}
\begin{quote}
\texttt{You are an intelligent chatbot designed for generating question-answer pairs from sentences.
}
\end{quote}
\noindent \texttt{USER}: \vspace{-0.7em}
\begin{quote}
\texttt{Your task is to generate a question and answer from the given sentence. \\
The question should start with 'Who'. \\
The question should refer to the subject of the given sentence.\\
The answer should include the subject of the given sentence. \\
Please generate the response in the form of a Python dictionary string with keys 'Q' for question and 'A' for answer. Each corresponding value should be the question and answer text respectively.\\
For example, your response should look like this: \{'Q': 'Your question here...', 'A': 'Your answer here...'\}.\\
Please note that the generated question and answer should only include information from the given sentence.\\
Please process the following sentence: \\
The man in the suit goes to the man in white and looks at him.
}
\end{quote}
\noindent \texttt{ASSISTANT}: \vspace{-0.7em}
\begin{quote}
\texttt{\{'Q': 'Who goes to the man in white?', 'A':'The man in the suit'\}
}
\end{quote}
\noindent \texttt{USER}: \vspace{-0.7em}
\begin{quote}
\texttt{Please process the following sentence: \\
<DECLARATIVE\_STATEMENT>
}
\end{quote}
\noindent \texttt{ASSISTANT}: \vspace{-0.7em}
\vspace{1em}

\subsection{Extracting Referring Expression Using Vicuna}
\label{subsection:grounding_ref_exp_rpompt_template}

In the quantitative evaluation, we use the following prompt template with Vicuna-13b-v1.5 to extract the referring expression from the output of the video-based LMM, which is used as the input prompt to the off-the-shelf-grounding module.

\vspace{0.5em}
\noindent \texttt{SYSTEM}: \vspace{-0.7em}
\begin{quote}
\texttt{
You are an intelligent chatbot designed for identifying the most relevant subject/object phrases in video-based question-sentence pairs. 
}
\end{quote}
\noindent \texttt{USER}: \vspace{-0.7em}
\begin{quote}
\texttt{Your task is to compare the question with the sentence, and extract the subject or object phrase of the sentence that most accurately answers the given question.\\
The selected phrase should be short and should contain only one noun.\\
The selected phrase can include adjectives that explain the attributes of the subject/object.\\
The selected phrase should not exceed 4 words.\\
The selected phrase should not include articles ('a', 'the', 'and').\\
Please generate the response in the form of a Python dictionary string with keys 'OBJECT', where its value is the extracted phrase in Python string format.\\
DO NOT PROVIDE ANY OTHER OUTPUT TEXT OR EXPLANATION. Only provide the Python dictionary.\\
For example, your response should look like this: \{'OBJECT': 'green toy'\}.\\
Please process the following video-based question-answer pair:\\
Question: who is in front of the guitar at the show? \\
Answer: A woman in a black dress is in front of the guitar on stage.
}
\end{quote}
\noindent \texttt{ASSISTANT}: \vspace{-0.7em}
\begin{quote}
\texttt{
\{'OBJECT': 'woman in black dress'\}
}
\end{quote}
\noindent \texttt{USER}: \vspace{-0.7em}
\begin{quote}
\texttt{
Question: who points to the window? \\
Answer: The old man is pointing to the window. \\
}
\end{quote}
\noindent \texttt{ASSISTANT}: \vspace{-0.7em}
\begin{quote}
\texttt{
\{'OBJECT': 'old man'\}
}
\end{quote}
\noindent \texttt{USER}: \vspace{-0.7em}
\begin{quote}
\texttt{
Question: who is inside the blue car? \\
Answer: The driver of the blue car. \\
}
\end{quote}
\noindent \texttt{ASSISTANT}: \vspace{-0.7em}
\begin{quote}
\texttt{
\{'OBJECT': 'driver'\}
}
\end{quote}
\noindent \texttt{USER}: \vspace{-0.7em}
\begin{quote}
\texttt{
Please process the following video-based question-answer pair:\\
Question: <INPUT\_TO\_VIDEO\_LMM> \\
Answer: <OUTPUT\_OF\_VIDEO\_LMM>
}
\end{quote}
\noindent \texttt{ASSISTANT}: \vspace{-0.7em}
\vspace{1em}

\subsection{Entity Matching with Vicuna}

As shown in Figure \ref{fig:overall-architecture}, our method employs an LLM-powered entity matching module similar to \cite{zhao2023bubogpt} to match the key phrases in the video-LMM's output with the object tracks obtained from the grounding and tracking module. We use the same prompt template as \cite{zhao2023bubogpt}.













\newpage
\onecolumn
\section{Qualitative Results for Visual Grounding}

\begin{figure*}[h]
   \centering
   \includegraphics[width=0.9\textwidth]{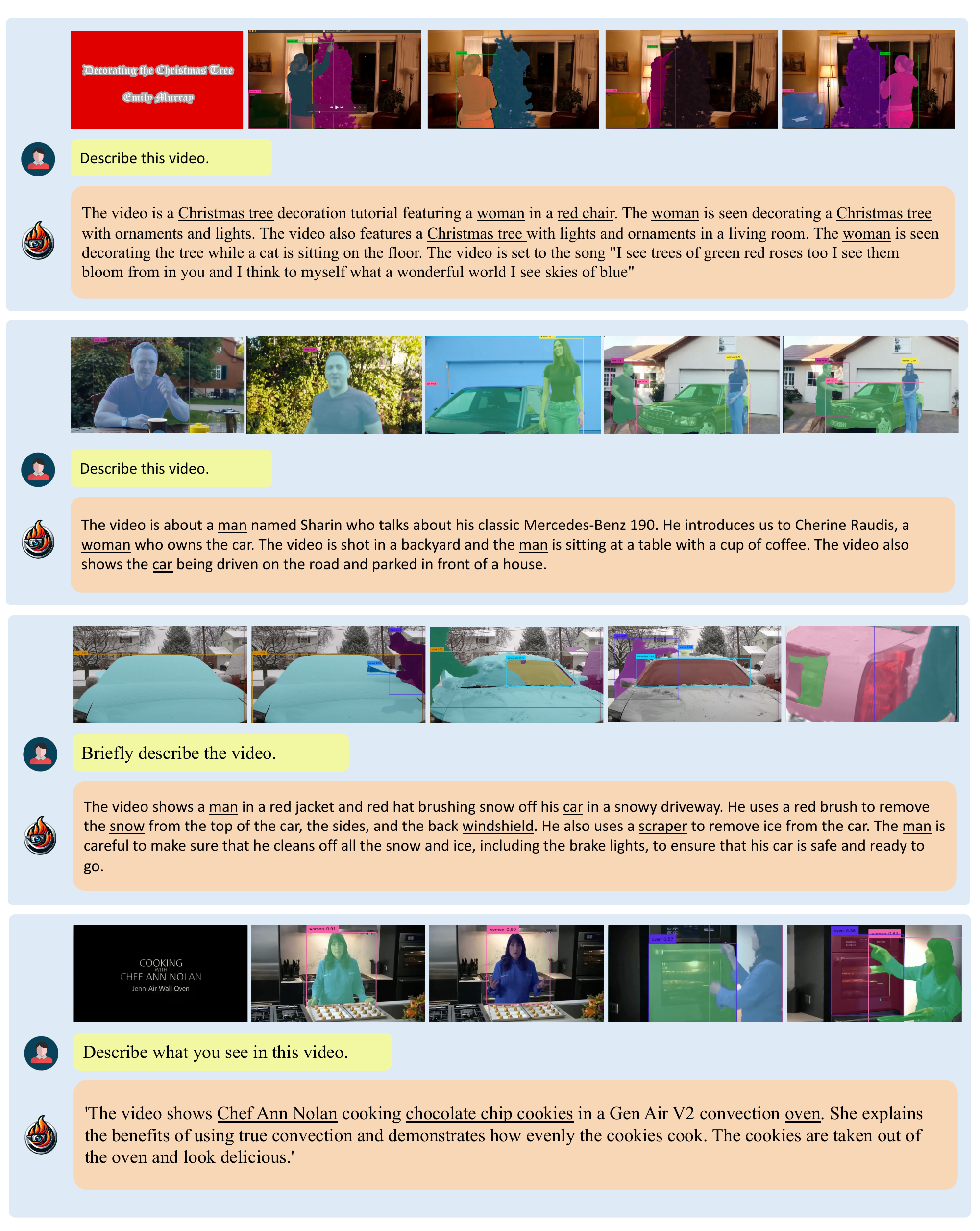}
   \caption{\textbf{Qualitative results for video grounding} obtained using image tags as the queries to the off-the-shelf grounding module and an entity matching module to match the image tags with the corresponding phrase in the LMM output. (e.g., in the 4th row, the phrase 'Chef Ann Nolan' is matched with the image tag 'woman'.)}
   \label{fig:qualitative-1}
\end{figure*}

\begin{figure*}[t]
   \centering
   \includegraphics[width=0.98\textwidth]{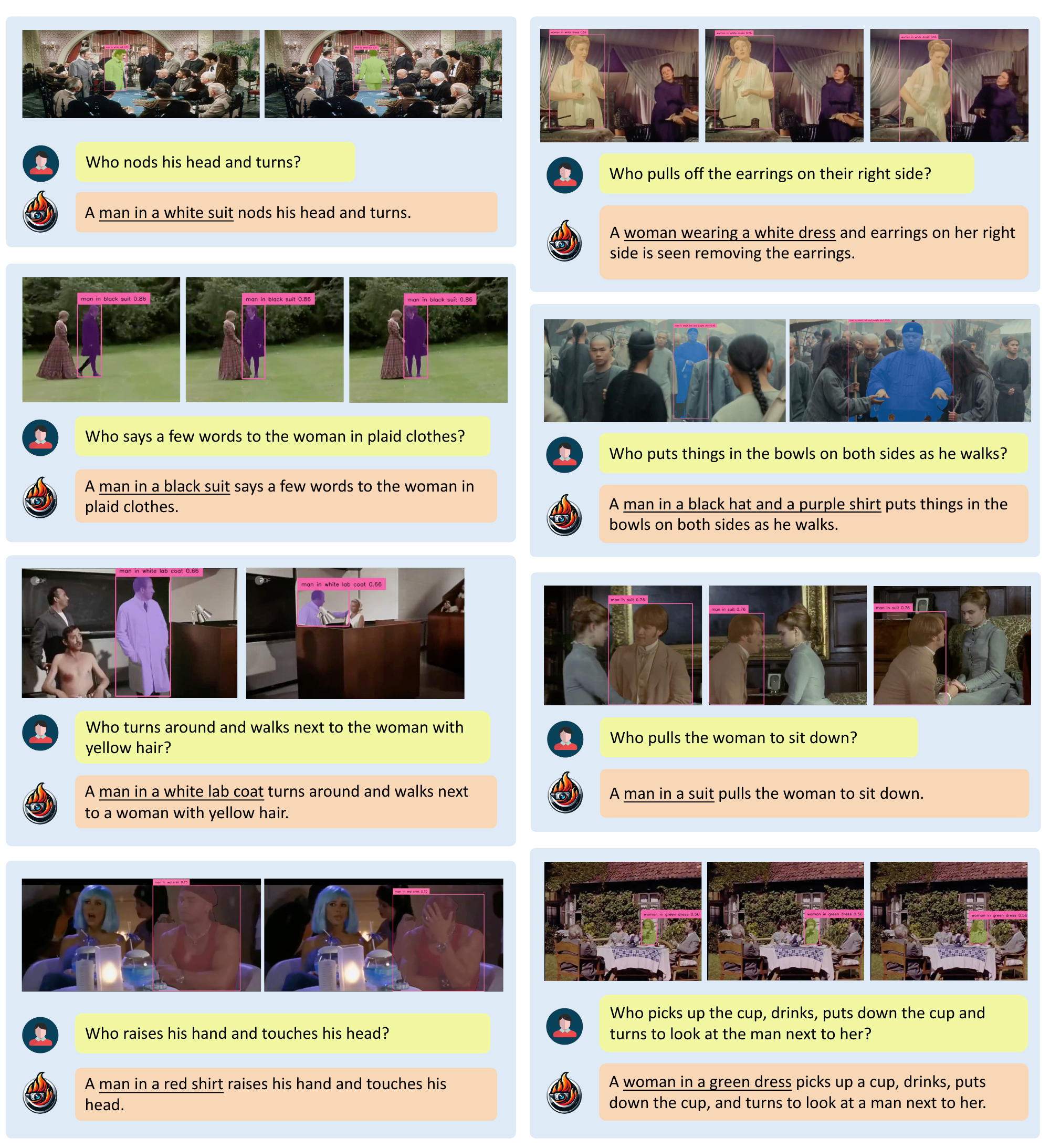}
   \caption{\textbf{Qualitative results for video grounding} on example videos from the HC-STVG\cite{hcstvg} dataset. These results are obtained by using Vicuna with the prompt template in \ref{subsection:grounding_ref_exp_rpompt_template} to extract the referring expression from the LMM output which is then passed to the off-the-shelf grounding module. }
   \label{fig:qualitative-2}
\end{figure*}

\begin{figure*}[h] 
   \centering 
   \includegraphics[width=0.98\textwidth]{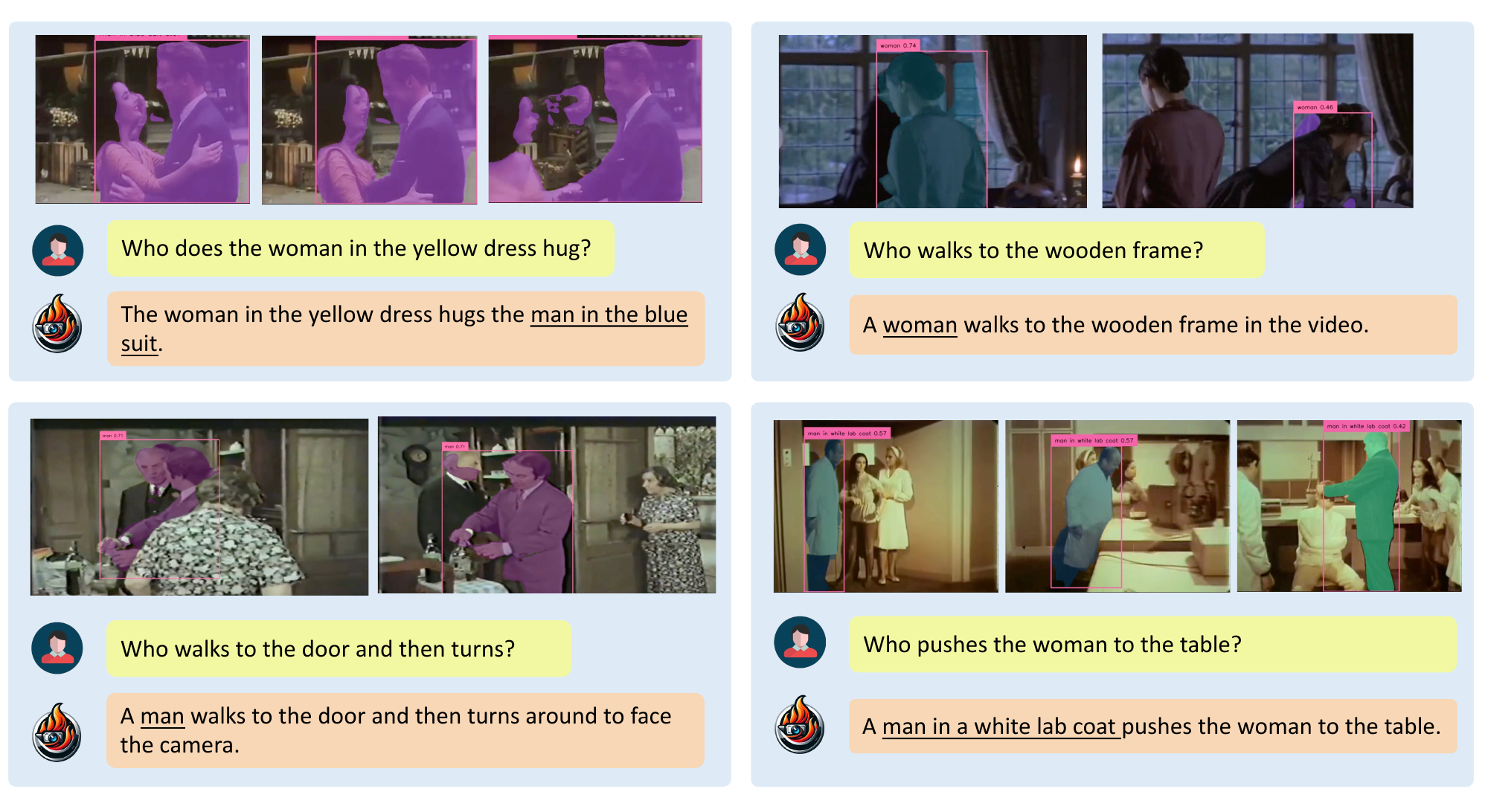} 
   \caption{\textbf{Qualitative results for visual grounding on the HC-STVG dataset (failure cases):} errors in our model's output (e.g., bottom-left: our model identifies the woman as a man), incorrect localizations in the off-the-shelf grounding module (e.g., top-left), and incorrect tracking (e.g., top-right, bottom-right) result in these failure cases.} 
   \label{fig:qualitative-4} 
\end{figure*}

\begin{figure*}[t]
   \centering
   \includegraphics[width=0.98\textwidth]{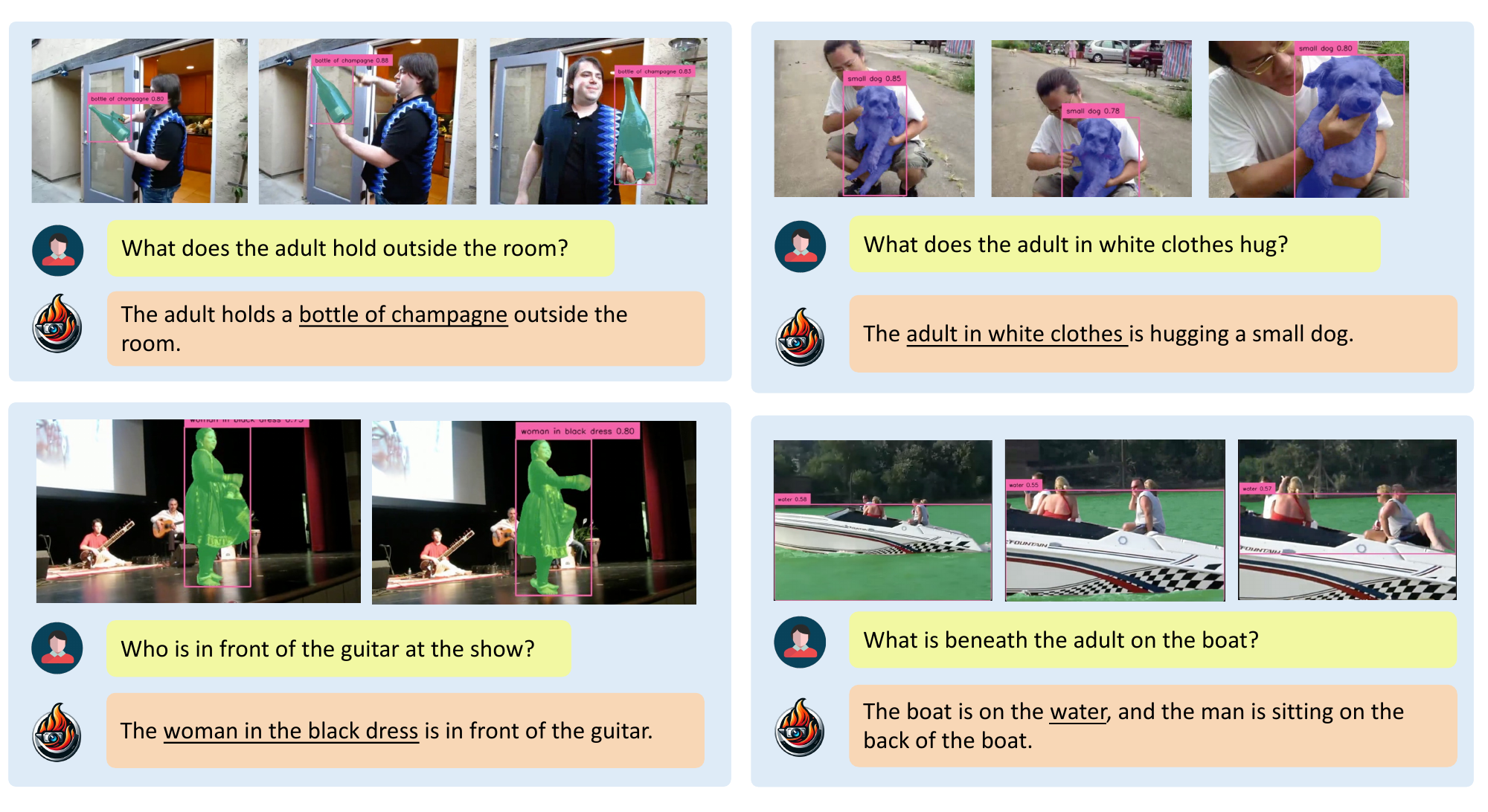}
   \caption{\textbf{Qualitative results for video grounding} on example videos from the VidSTG\cite{zhang2020does} dataset}
   \label{fig:qualitative-3}
\end{figure*}



\newpage
\twocolumn
\section{Quantitative Evaluations of Video-based Conversation Performance}

We leverage the video-based conversation performance benchmarks introduced in Video-ChatGPT\cite{Maaz2023VideoChatGPT}, while changing the evaluation LLM from GPT-3.5-Turbo to  Vicuna-13b-v1.5 model. The prompt templates used with Vicuna are as same as with \cite{Maaz2023VideoChatGPT}.

\noindent
\textbf{Video-based Generative Performance Benchmarking: } In this benchmark we continue to use the same test set of 500 samples curated from the ActivityNet-200\cite{Heilbron2015ActivityNetAL}
videos as in \cite{Maaz2023VideoChatGPT}.

\noindent
\textbf{Zero-Shot Question-Answer Evaluation: } Following Video-ChatGPT, we perform zero-shot evaluation on four standard open-ended question-answer datasets: MSRVTT\cite{xu2016msr-vtt}, MSVD\cite{xu2017video_msvdqa}, TGIF\cite{tgif-cvpr2016}, and ActivityNet-QA\cite{yu2019activityqa}. No specific training is performed on these datasets, and the evaluation is performed in a zero-shot manner.